\documentclass{svproc}
\usepackage{comment}
\usepackage{graphicx}
\usepackage{hyperref}
\usepackage{cleveref}

\begin{document}
\mainmatter              
\title{An Interpretable Systematic Review of Machine Learning Models for Predictive Maintenance of Aircraft Engine}
\author{Abdullah Al Hasib \and Ashikur Rahman \and
Mahpara Khabir \and Md. Tanvir Rouf Shawon}
\titlerunning{Predictive Maintenance of
Aircraft Engine}  

\authorrunning{ Abdullah Al Hasib et al.}

\institute{Department of Computer Science and Engineering, Ahsanullah University of Science and Technology, Dhaka, Bangladesh\\
\email{aahasib.aust@gmail.com, ashiq4998@gmail.com, khabirmk98@gmail.com, shawontanvir95@gmail.com}}

\maketitle              
\begin{abstract}
This paper presents an interpretable review of various machine learning and deep learning models to predict the maintenance of aircraft engine to avoid any kind of disaster. One of the advantages of the strategy is that it can work with modest datasets. In this study, sensor data is utilized to predict aircraft engine failure within a predetermined number of cycles using LSTM, Bi-LSTM, RNN, Bi-RNN GRU, Random Forest, KNN, Naive Bayes, and Gradient Boosting. We explain how deep learning and machine learning can be used to generate predictions in predictive maintenance using a straightforward scenario with just one data source. We applied lime to the models to help us understand why machine learning models did not perform well than deep learning models. An extensive analysis of the model's behavior is presented for several test data to understand the black box scenario of the models. A lucrative accuracy of 97.8\%, 97.14\%, and 96.42\% are achieved by GRU, Bi-LSTM, and LSTM respectively which denotes the capability of the models to predict maintenance at an early stage.
\end{abstract}

\section{Introduction}
It is true that air travel is one of the safest ways of transportation. However, when they do occur, aviation accidents frequently have disastrous consequences. approximately the course of a recent five-year period, there were approximately 4,000 accidents\footnote{\url{https://www.psbr.law/aviation_accident_statistics.html}} with engine failure as a contributing factor, or nearly two accidents every day. the NTSB claims. Indian Navy reports the fifth MiG-29K crash in four years\footnote{\url{https://www.janes.com/defence-news/news-detail/indian-navy-reports-crash-of-fifth-mig-29k-in-four-years}}

A piece of machinery’s capacity cannot be maintained indefinitely; occasionally,
it will break down due to antiquated procedures. Systems for monitoring machinery
that incorporate sensors can only report on the state of the equipment
and not whether it is in good or bad condition. An inspection is performed on
a machine to prevent the worst scenario (failure) and to learn more about its
condition. To the scheduled machinery system, strategy must be used. Three
maintenance strategies represent the best practices: Predictive and preventive
maintenance Continual preventive maintenance is the foundation of predictive
maintenance (PdM). keeping an eye on the machine’s condition and performing
maintenance as best you can. Based on historical data, integrity factors, statistical inference techniques, and engineering methodologies, PdM indicated the machine’s state for scheduling maintenance.
Predictive modeling research has advanced significantly in recent years, both academically and commercially. Predictive maintenance modeling techniques come in four different flavors: knowledge-based, data-driven, physics-based, and hybrid-based \cite{Ref1}\cite{Ref2}.

In aircraft, predictive maintenance is essential for lowering costs, improving safety, and raising reliability. It improves maintenance expenditures and reduces downtime by identifying probable issues early. By averting crises during flight, this proactive method guarantees passenger safety. Additionally, it increases the longevity of assets and makes data-driven scheduling decisions possible. Predictive maintenance transforms maintenance procedures by utilizing cutting-edge analytics and real-time monitoring, increasing operational effectiveness and efficiency. It converts conventional reactive maintenance into a proactive, preventative method, having a considerable positive impact on airlines and the aviation sector as a whole. Several machine learning and deep learning models like LSTM, Bi-LSTM, RNN, Bi-RNN GRU, Random Forest, KNN, Naive Bayes, and Gradient boosting have been examined here in this task in search of a reliable model. In an earlier study \cite{Ref3}, researchers used Random Forest, and they were successful with lower  score accuracy. However, this model allowed us to attain maxmimum accuracy in our research. Another study \cite{Ref5} \cite{Ref6} employed LSTM, but with an accuracy rate of 87.3\% and 96\%. However, LSTM achieved better accuracy in our situation. As a result, we outperform our work in both instances. Another issue is that, while we utilize lime to categorize samples based on feature values, no researchers ever show how to do so.

In our work, we have tried to build a reliable model for the predictive maintenance of aircraft engine from sensor data. A very well-known dataset \textbf{CMAPSS} by NASA is used in our work after doing some necessary preprocessing. 
Our contribution can be summarized in the following points:
\begin{itemize}
    \item  We have used a very well-known time series dataset (CMAPSS) in our work to do the task of predictive maintenance of aircraft engine. Using a simple scenario and only one data source (sensor readings), we demonstrate how deep learning and machine learning can be used to create predictions in predictive maintenance.
    \item  LSTM, Bi-LSTM, RNN, Bi-RNN GRU, Random Forest, KNN, Naive Bayes, and Gradient boosting are used to use sensor information to predict aircraft engine failure within a specified number of cycles. 
    \item We have also evaluated our models with several performance metrics like accuracy, precision, recall, and f1 score and explained the predictions of the models with an explainable AI technique LIME.
\end{itemize}

\section{Related Work}

A work \cite{Ref3} by Adryan Fitra Azyus et al. employed both classification and regression algorithms. The LSTM method’s
98.7\% accuracy, 92.3\% precision, and 96.6\% recall rate
are the greatest. The regression strategy for predictive maintenance works best when the Random Forest method is used and accuracy is achieved at 90.3\%. A work \cite{Ref4} by Adryan Fitra Azyus et al. building many GRU units allowed them to test out the GRU architecture in this study. Although the gru1 model best fits the FD001 dataset, RMSE is 20.5502224 and MSE is 427.197998.

In this paper \cite{Ref5} by Srikanth Namudur et al. five algorithms in total were put into practice and evaluated. With an accuracy of 87.3 percent, LSTM can accurately identify 268 out of 307 mistakes. This indicates the superiority of LSTM for sequential data learning. In another paper \cite{Ref6} by Abdeltif Boujamza et al. an aircraft engine's remaining life can be predicted with up to 96\% accuracy using LSTM. Circular connections in the LSTM allow it to investigate the sequential dynamics of sensor data in hypothetically predicative situations.

In \cite{Ref7} by Khalid Khan et al. the trend for predicting aviation engine and hydraulic system failure is trending towards neural networks, specifically LSTM, given the time series nature of the aircraft engine data. In this work \cite{Ref8} it acts as a state-of-the-art review to pinpoint the cutting-edge approaches being used to tackle PdM issues and map the present state of the field. In this paper \cite{Ref9} by Maren David Dangut et al. the AE-BGRU and AE-CNN-BGRU, two bidirectional models, were taken into account and compared with GRU. According to the evaluation's findings, the AE-CNN-BGRU model is able to manage irregular patterns and trends efficiently, which helps to solve the imbalanced classification issue.
In another research \cite{Ref10}, a cutting-edge method for failure diagnosis based on the equipment degradation sequence and recurrent neural networks was proposed. Due to its ability to learn long-term dependencies, the LSTM network is used in this situation. Research \cite{Ref11} by Ingeborg de Pater, Arthur Reijns, and Mihaela Mitici suggested a dynamic maintenance structure where component RUL prognostics are routinely updated. Long-term statistics indicate that only 7.4\% of the overall maintenance expenditures are attributable to engine problems. The maintenance expenses are 24.4\% greater as compared to the ideal scenario with flawless RUL prognostics.

In \cite{Ref12}, the most modern design and optimization machine learning algorithms for aviation maintenance are thoroughly reviewed and compared analytically. Various evaluation parameters, such as time and cost of maintenance, are explained and summarized in order to optimize the aviation maintenance system.
In this paper \cite{Ref13} compared to mean-estimated-RUL's maintenance scheduling, our DRL strategy lowers overall maintenance expenses by 29.3\%. Additionally, it avoids 95.6\% of unexpected engine replacements. The engines have an average wasted life of just 12.8 cycles before they are replaced.

\section{Dataset}
The dataset is put together by the NASA Ames Prognostics CoE, which is called CMAPSS\footnote{https://data.nasa.gov/dataset/C-MAPSS-Aircraft-Engine-Simulator-Data/xaut-bemq} Dataset. The dataset consists of several multivariate time series. Three operational parameters have a considerable impact on the engine's performance. The dataset consists of 26 columns of numbers, each separated by a space. Every row denotes a snapshot of data that was captured during an operational cycle. A separate variable appears in each column. It is important to forecast when an operational engine will fail based on aircraft engine operating data and failure event history. The feature details can be seen in Table \ref{tab:table1}.

\begin{table}[htb]
\caption{Feature details of CMAPSS Dataset}
\centering
\label{tab:table1}
\begin{tabular}{ c|c } \hline
\textbf{Feature Name} &  \textbf{Feature Details} \\ \hline
ID & Machine Id  \\ 
Cycle & Time Step \\ 
Setting1, Settting2 and Setting3  & Operational Settings\\  
S1 - S21 & Sensors measurement \\ \hline
\end{tabular}
\end{table}

\section{Proposed Methodology}
This section covers the entire methodology, along with proposed models and the methods for processing the data.

\subsection{Data Preprocessing}
The four sets of data in the CMAPSS Dataset are designated as FD001, FD002, FD003, and FD004, and each is broken down into Train, Test, and RUL data. The initial step is to create labels for the training data. These labels are rul (remaining usable life), Label1, and Label2, much like in the predictive maintenance template.
\begin{figure}[h]
\centering
\includegraphics[width=90mm,height=8cm]{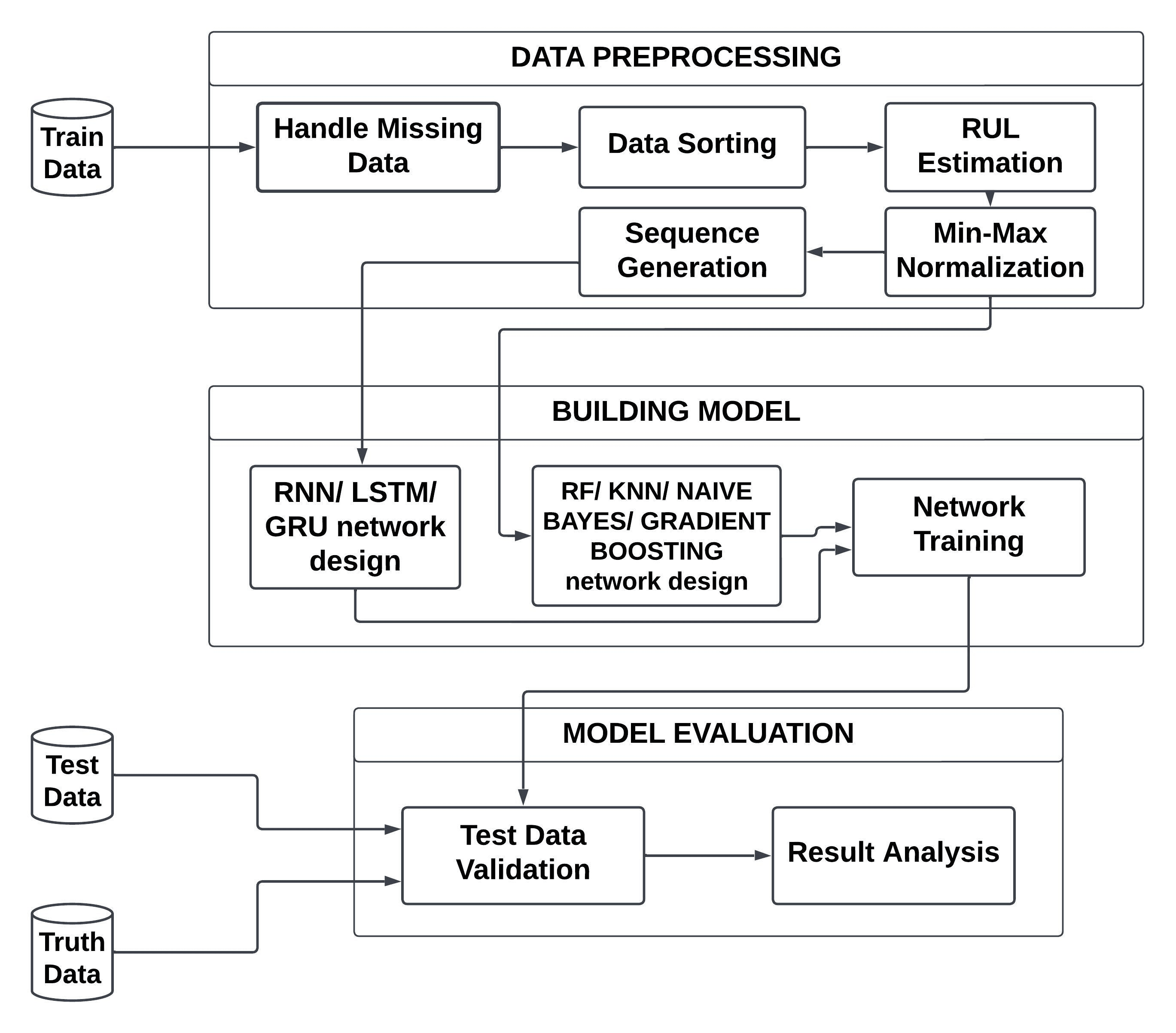}
\caption{Methodology of our proposed work}
\label{fig:methodology}
\end{figure}
When calculating the RUL value for each cycle, the highest value of the cycle is subtracted from each cycle before being added to the RUL value for each machine id.
Equation \ref{eq-bnhtr} looks like this:

\begin{equation} \label{eq-bnhtr}
RUL = RUL_(true\_from\_dataset) + RUL_(cycle\_max-cycle) 
\vspace{2mm}
\end{equation} 

In order to manually estimate each machine's maximum life and the point of deterioration (reduction in machine performance), each machine id has a distinct number of cycles. The algorithm is as follows:
\\
if n in $RUL \geq $Maxlife : newRUL = Maxlife else newRUL = n 
\\
We adjust the amount of the data dimensions while processing data to become RNN, LSTM, or GRU input since we are aware that deep learning requires high-dimension input data, particularly for LSTM and GRU. We must perform a dimensional transformation to a higher level in order to allow the artificial neural network, specifically the LSTM, GRU that was constructed, to incorporate more feature levels. By estimating the size of the data sequence to be 50, the process of converting this dimension from 2 dimensions to 3 dimensions is carried out. As a result, a two-dimensional data size of (50, 26) will be created, and this value can be used to choose the batch size for the training procedure. The overall methodology is shown in Fig. \ref{fig:methodology}
\newline For machine learning models, feature-wise data will be delivered to the Random Forest, KNN, Naive Bayes, and Gradient Boosting networks following min-max normalization in order to train the model.

\subsection{Models}
We have built both machine learning and deep learning architectures, and a comparison of their performance is also discussed, to forecast airplane engine failure within a predetermined number of cycles.

\textbf{• Machine Learning Models:} An ensemble machine learning technique called \textbf{Random Forest (RF)} is well-suited for high-dimensional datasets, noisy data, and missing values, and can provide feature importance rankings. 
An efficient machine learning approach called \textbf{K-Nearest Neighbors (KNN)} predicts the class or value of an input sample based on the dominant class.
\textbf{Naive Bayes }Classifier is a simple powerful probabilistic machine learning algorithm used for classification tasks, where it predicts the class of an input sample based on Bayes' theorem. 
\textbf{Gradient Boosting }Classifier is used for classification tasks, where it combines multiple weak decision trees to create a strong classifier.

Since it is well known that lime cannot be utilized for merged data, machine learning models are used in this case as well as to see if they perform well or worse than deep models. The hyperparameters are also modified to produce higher accuracy than in earlier studies.

\textbf{• Deep Learning Models:} The outputs of one step are utilized as inputs for the subsequent step in a \textbf{Recurrent Neural Network (RNN)}. RNNs' Hidden state, which retains certain sequence-related information, is its primary and most significant property. The hidden state for each time step is simultaneously determined by \textbf{Bidirectional RNNs} using data from previous and succeeding time steps.  
\textbf{LSTM }is a type of recurrent neural network (RNN) that is designed to handle the vanishing gradient problem, which can occur in traditional RNNs when training on long sequences of data.  \textbf{Bidirectional long-short term memory (bi-lstm)}, which enables any neural network to retain sequence information in both ways (backward, from future to past, or forward), is a word used to describe this (past to future). 
\textbf{GRU} is increasingly being used in deep learning models for time series analysis. In a GRU, the cell states and hidden states are merged with the input gate and forget gate of an LSTM as well as the update gate. 

According to past research \cite{Ref5} \cite{Ref6}, deep learning models work effectively. For this reason, in the research of predictive maintenance, we are trying to improve upon those models.

\subsection{Evaluation of the models} Our models are evaluated using accuracy, precision, recall, and F1-score. How often the model was overall correct may be determined by its accuracy. How well the model can forecast a certain category is determined by its precision. The recall reveals how frequently the model was able to recognize a certain category. The F1-score combines accuracy and recall from a classifier into a single statistic by computing the harmonic mean of these two metrics.

\section{Model Architecture}
The rolling function of the predictive maintenance template can be calculated in a manner similar to this by selecting a window size of 50 cycles \cite{Ref14}. \textbf{LSTM} is intended to replace the need for manual model development by enabling the extraction of abstract characteristics from a set of sensor values in a window. The first layer is a 100-unit LSTM layer, while the subsequent layer is a 50-unit LSTM layer. The total parameter count for LSTM is shown in Table \ref{tab:table2}.
 A \textbf{RNN} network is constructed. Following, the relu activation function is the first layer, a RNN layer with 32 units. The relu activation function is employed to lessen the issue of vanishing gradients in the first eight-unit dense layer \cite{Ref15}. Table \ref{tab:table2} displays the total parameter of RNN.
 Every time the 3-dimensional input layer (none, 50, 100) passes through the \textbf{GRU} layer and the Fully Connected layer, it is activated using a sigmoid function to generate an output in the form of a single number, the RUL value. In terms of the experiment, a total of four GRU architectures 1 GRU through 4 GRU were tested. Table \ref{tab:table2} displays the total parameter information.
 
Some machine learning models are also used. Here \textbf{Random Forest} and \textbf{Naive Bayes} were imported from sklearn. \textbf{KNN} was also imported from sklearn and it's neighbor size is 3. 
In \textbf{Gradient boosting} the max features are 2 and the max depth is 2.

\renewcommand{\arraystretch}{2}
\begin{table}[h]
\caption{Total Parameter Count of the deep learning models}
    \label{tab:table2}
    \centering
    \begin{tabular}{p{15mm}|p{55mm}} \hline
       \textbf{Model} &  \textbf{Total Trainable Parameter}  \\  \hline
       LSTM & 76651  \\ 
       RNN & 657   \\ 
       GRU & 113301   \\ \hline
    \end{tabular}
\end{table}
\section{Experiments and Result Analysis}
This section shows the experimental results, analysis of the models and explanation of the models using the interpretable technique - LIME in 3 separate subsections.
\subsection{Experimental Setup:}
CMAPSS Dataset was run through the previously mentioned machine learning and deep learning algorithms. Our datasets were split into train and test sets containing 80\% and 20\%. We used the sklearn package to use the algorithms. We used a combination of Python 3.6.5, Keras 2.3.1 and Tensorflow 2.0.0. For the machine learning model's performance evaluation, we used Lime also. Google Colaboratory was used to implement our codes. In \textbf{RNN} the relu activation function is used in the first layer and the sigmoid activation function is used in the last layer. We employed a batch size of 64, an epoch of 15 for the training parameters, and a learning rate with an Adam optimizer of 0.001. Here, the validation split is 0.1 and verbose=1. The early stopping method is also used.
In \textbf{LSTM} relu activation function is used in the first layer and softmax is used in the last layer. The early stopping method is also used here. We employed a batch size of 64, an epoch of 15, a dropout of 0.5, and the learning rate is 0.001. Here the validation split is 20\% and verbose=0. 
Theoretically, the learning process will be deeper and the reach of feature levels will be larger the deeper the network is built. In \textbf{GRU} we employed a batch size of 200, a dropout of 0.2, a learning rate with an Adam optimizer of 0.001, and epochs of 20 for the training parameters investigated in this work.

We import \textbf{Random Forest} from sklearn.
In \textbf{KNN} we import KNeighborsClassifier with neighbors size=3 from sklearn. 
We also import \textbf{Naive Bayes} from sklearn.
In \textbf{Gradient Boosting}, the estimators are 20, and the learning rate is 0.5.

\subsection{Experimental Result}
Results achieved by our models can be seen in this section having 2 different tables along with the analysis of the results. Table \ref{tab:table3} shows the results of deep learning models and table \ref{tab:table4} shows the performance of machine learning models.

\renewcommand{\arraystretch}{2}
\begin{table}[h!]
\caption{Performance Metrics (Deep Learning Models)}
\centering
    \label{tab:table3}
    \begin{tabular}{p{25mm}|p{25mm}|p{18mm}|p{16mm}|p{20mm}} \hline
      \textbf{DL Models} &  \textbf{Accuracy(\%)} &  \textbf{Precision}&  \textbf{Recall}&
      \hspace{.2cm}\textbf{F1-score} \\ \hline
       LSTM &   96.42 &   0.10 &   0.93 &   0.96\\ 
        RNN &  91.40 &  0.99 &  0.96 &  0.97\\ 
       Bi-LSTM &  97.14 &  0.10 &  0.94 &  0.97\\ 
      Bi-RNN & 94.85 & 0.10 & 0.96 & 0.98\\ 
       GRU &  97.80 &  0.93 &  0.10 &  0.96\\ \hline
    \end{tabular}
\end{table}

\renewcommand{\arraystretch}{2}
\begin{table}[h!]
  \caption{Performance Metrics (Machine Learning Models)}
  \centering
    \label{tab:table4}
    \begin{tabular}{p{32mm}|p{25mm}|p{18mm}|p{16mm}|p{20mm}} \hline
      \textbf{ML Models} &  \textbf{Accuracy(\%)} &  \textbf{Precision}&  \textbf{Recall}&
      \hspace{.2cm}\textbf{F1-score} \\ \hline
      
       Random Forest &  96.1\% &  0.89 &  0.83 &  0.86\\ 
       KNN &  97.3\% &  0.93 &  0.88 &  0.90\\ 
       Naive Bayes &  93.0\% &  0.68 &  0.96 &  0.80\\ 
       Gradient Boosting &  95.6\% &  0.87 &  0.81 &  0.84\\ \hline

    \end{tabular}
\end{table}

\begin{figure}[h]
\centering
    \includegraphics[width=0.4\textwidth]{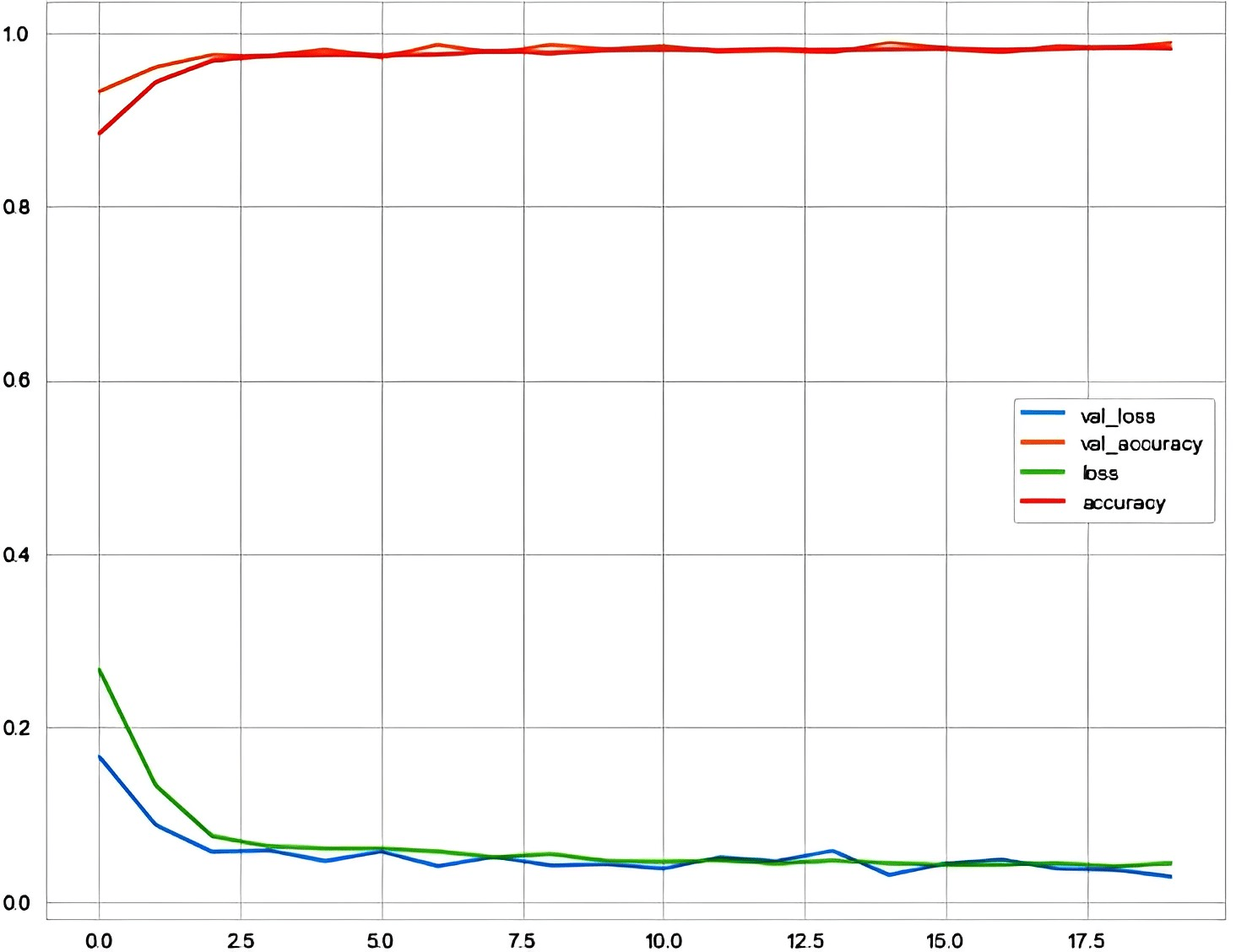}
    \caption{Loss and accuracy regarding the epochs of GRU (Best performing model)}
    \label{fig:gru}
\end{figure}

Fig. \ref{fig:gru} shows the loss and accuracy curve of GRU. The loss and accuracy curve for a GRU model trained for 20 epochs shows metrics such as validation loss, validation accuracy, accuracy, and loss. These metrics offer information on the model's capacity for generalizing what it has learned from the training set of data to new, untried data. Overfitting of the model may be revealed if the validation loss starts to rise while the training loss keeps falling. Due to repeated Jacobian multiplications during backpropagation, RNNs have gradient vanishing and explosion issues. RNNs are unable to simulate long-term sequential dependencies because the gradient signal used for training is inconsistent. Compared to RNN, Bi-RNN networks have better accuracy. Long short-term memory networks (LSTM) prevent vanishing gradients by updating cell states in an additive, non-multiplicative manner. However, it can suffer from gradient explosions. Bi-LSTM also performs well and achieved the second-highest accuracy because the input flows in both directions and it is able to take data from both sides.

\subsection{Result analysis}
In their iterative layers, feedback loops are a feature of all RNNs. As a result, information can be kept in "memory" for a long time. It can be challenging to train a normal RNN to handle issues that call for learning long-term temporal dependencies, though. This is due to the loss function's gradient's exponential decrease over time (called the vanishing gradient problem). A particular kind of RNN called an LSTM network employs both unique and conventional units. The timing of when data is output, stored in memory, and forgotten is controlled by a sequence of gates. Long-term dependencies can be learned using this design. GRU employs a structure that is simpler than LSTM but is similar to it. Although it does not require individual memory cells and makes use of fewer gates, it nevertheless uses a number of gates to regulate the information flow.

Long short-term memory (LSTM) training outcomes outperform recurrent neural network (RNN) training results, according to a comparison of the two. First, a simple RNN is used with two dense layers to get the output with 91.40\% accuracy shown in Table \ref{tab:table3}.   However, compared to RNN, which has an accuracy of 94.85\%, Bi-RNN networks have better accuracy. Strategies such as gradient clipping are often used to mitigate this problem. So a 2-layer LSTM is used and gets 96.42\% accuracy. And after using Bi-LSTM, the maximum accuracy has been discovered, which is 97.14\%.  The GRU continues to outperform other algorithms according to the total comparative data. The accuracy that has been achieved in this experiment was 97.80\%, which is significantly higher than that of any other models. The benefits of employing GRU to forecast the projected value of RUL are in terms of processing effectiveness and research expenditure expenses since the RUL approach can be processed even with a medium-spec PC, making machine learning and deep learning processing more affordable in the future. input flows in both directions and it is able to take data from both sides. 
\\
Data are merged during training by using deep learning models LSTM/ RNN/ GRU, which is a common occurrence for sequential modeling, and the results that have been found are shown earlier. But while applying machine learning models, feature-specific data are used and there is no data merging. The accuracy for Random Forest, KNN, Naive Bayes, and Gradient Boosting are 96.1\%, 97.3\%, 93.0\%, and 95.6\% respectively in Table \ref{tab:table4}. But the corresponding Precision, Recall, and F1-score values are lower than the other deep learning models. Compared to deep learning models, machine learning models performed comparably lower. The Lime technique has been used to machine learning models to explain why they didn't perform well.

\subsection{Explanation of the models through LIME} 
\begin{figure}[h]
    \centering
    \includegraphics[width=0.65\textwidth]{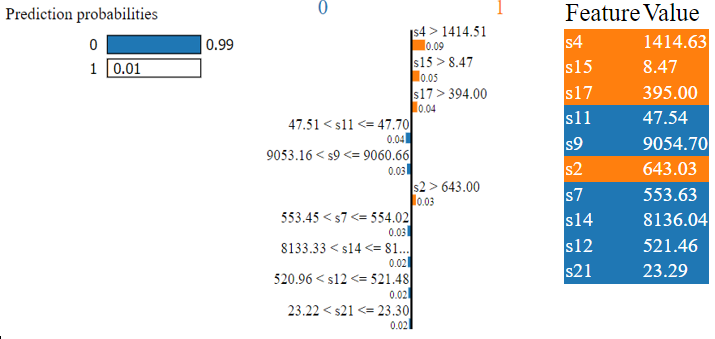}
    \caption{True Positive Case of RF Classifier}
    \label{fig:sample6RF}
\end{figure}

\begin{figure}[h]
    \centering
    \includegraphics[width=0.65\textwidth]{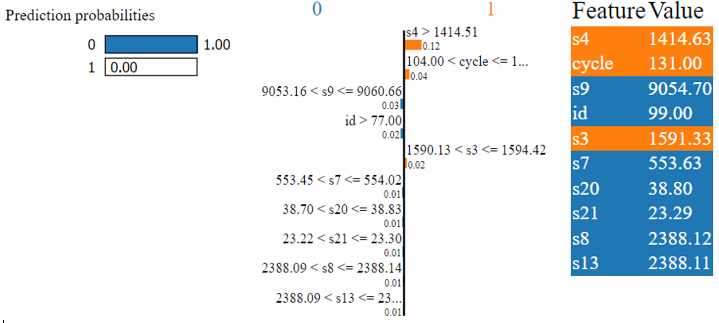}
    \caption{True Positive Case of KNN Classifier}
    \label{fig:sample6RNN}
\end{figure}

\begin{figure}[h]
    \centering
    \includegraphics[width=0.65\textwidth]{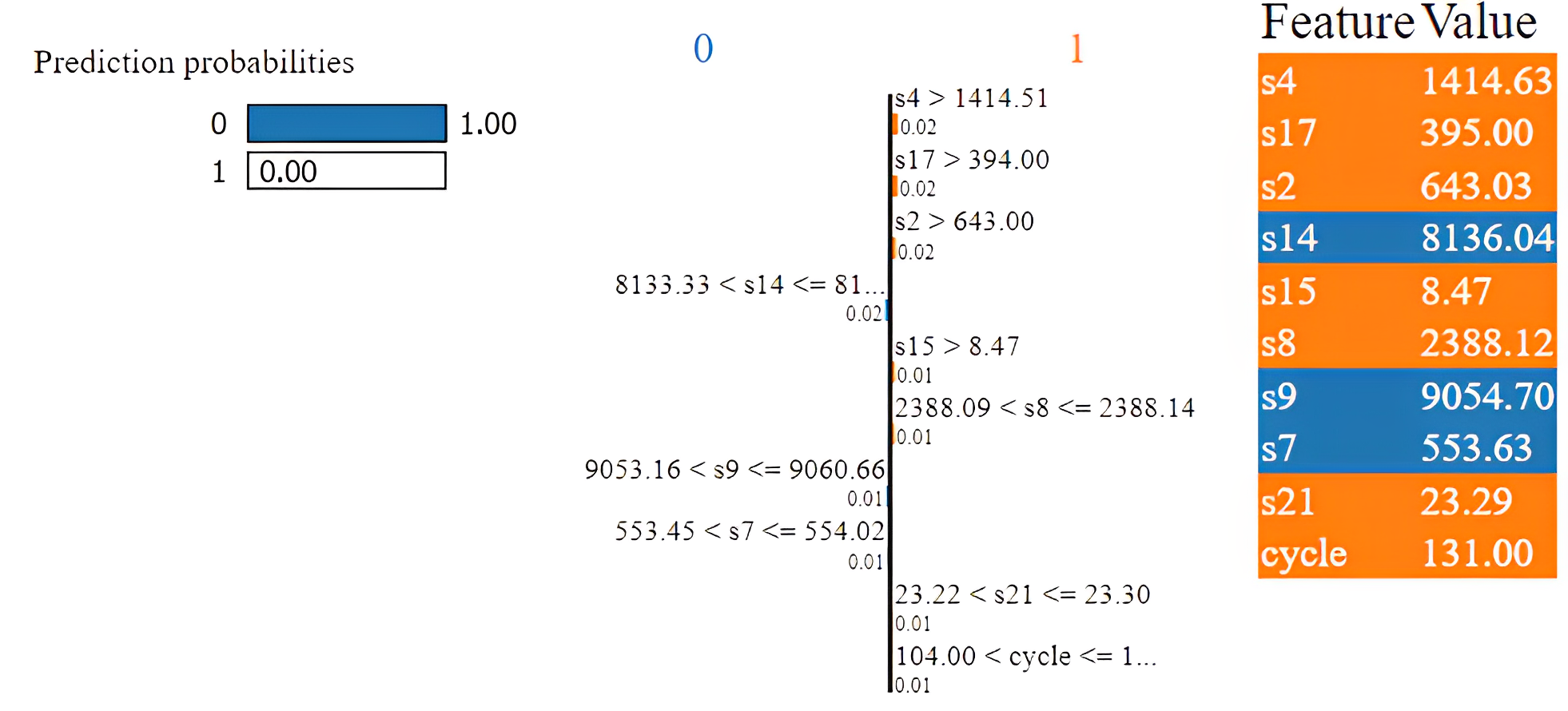}
    \caption{True Positive Case of Naive Bayes Classifier}
    \label{fig:samp6_nb_final}
\end{figure}

\begin{figure}[h]
    \centering
    \includegraphics[width=0.65\textwidth]{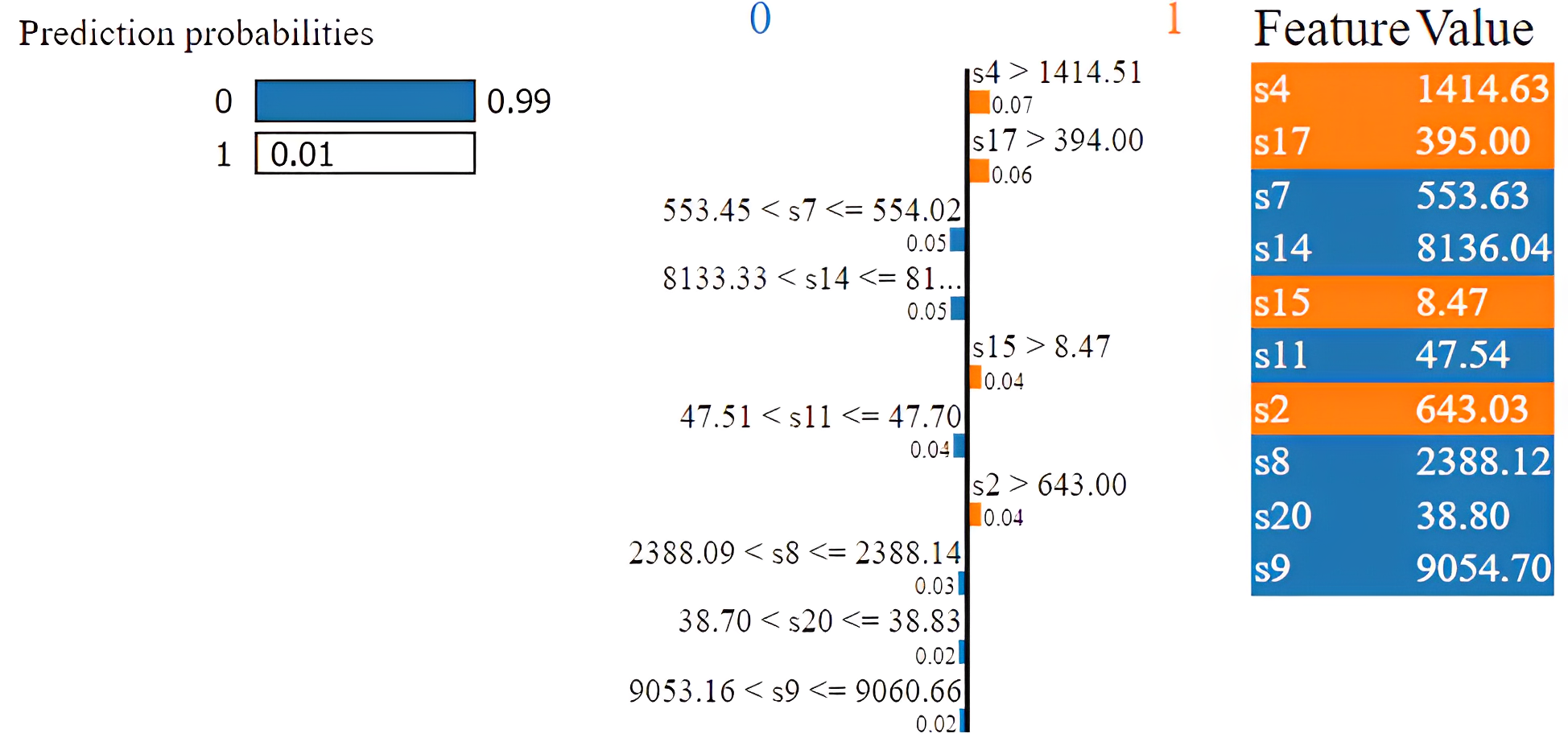}
    \caption{True Positive Case of Gradient Boosting Classifier}
    \label{fig:samp6_gradboost_final}
\end{figure}

The method known as Local Interpretable Model-agnostic Explanations (LIME) is used frequently to explain predictions made by black-box machine learning models. To explain specific predictions made by black box machine learning models, local surrogate models, which are interpretable, are used. Text classifiers, tabular models, and image classifiers may all be explained using LIME. In order to demonstrate how each sample is categorized depending on the values of the features, we implemented machine learning models to our dataset, including Random Forest, KNN, Naive Bias, and Gradient Boosting, and then applied lime to it.
\\
We chose sample number 6 and applied lime on it after we trained our dataset using the four machine-learning models in the order indicated above. In the test set, the maintenance classified as 0 is represented by the sixth row. It defines, there is no maintenance required, and it was correctly classified by all models. Fig. \ref{fig:sample6RF} shows that the Random Forest model is 99\% confident that it is under the label of 0 classification. The values of s11, s9, s12, s14, and s7 increase maintenance chance to be classified as 0.
According to Fig. \ref{fig:sample6RNN}, the KNN classifier is absolutely confident that it is classed as 0, and the values of id and s9 increase the likelihood that it will be given that classification.
Fig. \ref{fig:samp6_nb_final} shows that the Naive Bayes model is 100\% guaranteed that the given sample belongs in the category of 0. The likelihood of it being classed as 0 grows with the values of s14, s9, and s7.  Once more, Fig. \ref{fig:samp6_gradboost_final} shows that the Gradient Boosting model is 99 percent confident that the provided sample will be categorized as 0, which is completely correct. This is due to the higher values of s7, s14, s11, s8, s20, and s9. 

\begin{figure}[h]
    \centering
    \includegraphics[width=0.65\textwidth]{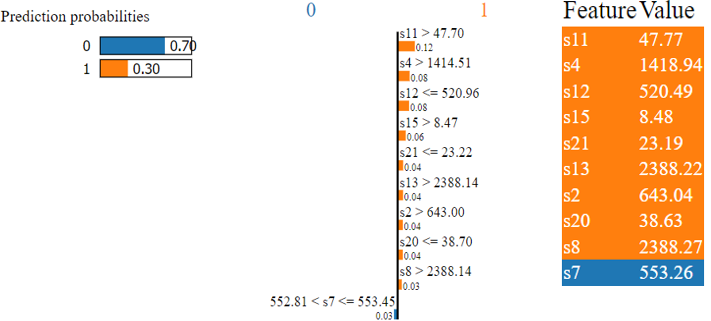}
    \caption{False Positive Case of Random Forest Classifier}
    \label{fig:sample4RF}
\end{figure}

\begin{figure}[h]
    \centering
    \includegraphics[width=0.65\textwidth]{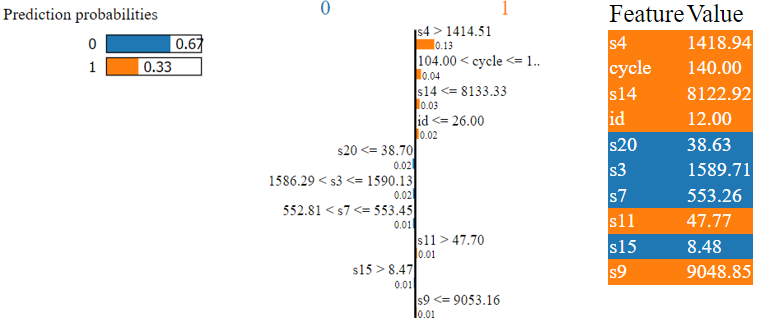}
    \caption{False Positive Case of KNN Classifier}
    \label{fig:sample4RNN}
\end{figure}

Following that, we chose sample number 4, whose real label reads 1, indicating that maintenance is required. In this instance, the Random Forest model has a 30\% confidence level that is categorized as 1. S7 values reduce the chance of correctly classifying it. The outcome is shown in Fig. \ref{fig:sample4RF}. Again, based on Fig. \ref{fig:sample4RNN}, we can see that the KNN classifier misclassified it for the values of s9, s3, and s21.

\begin{figure}[h]
    \centering
    \includegraphics[width=0.65\textwidth]{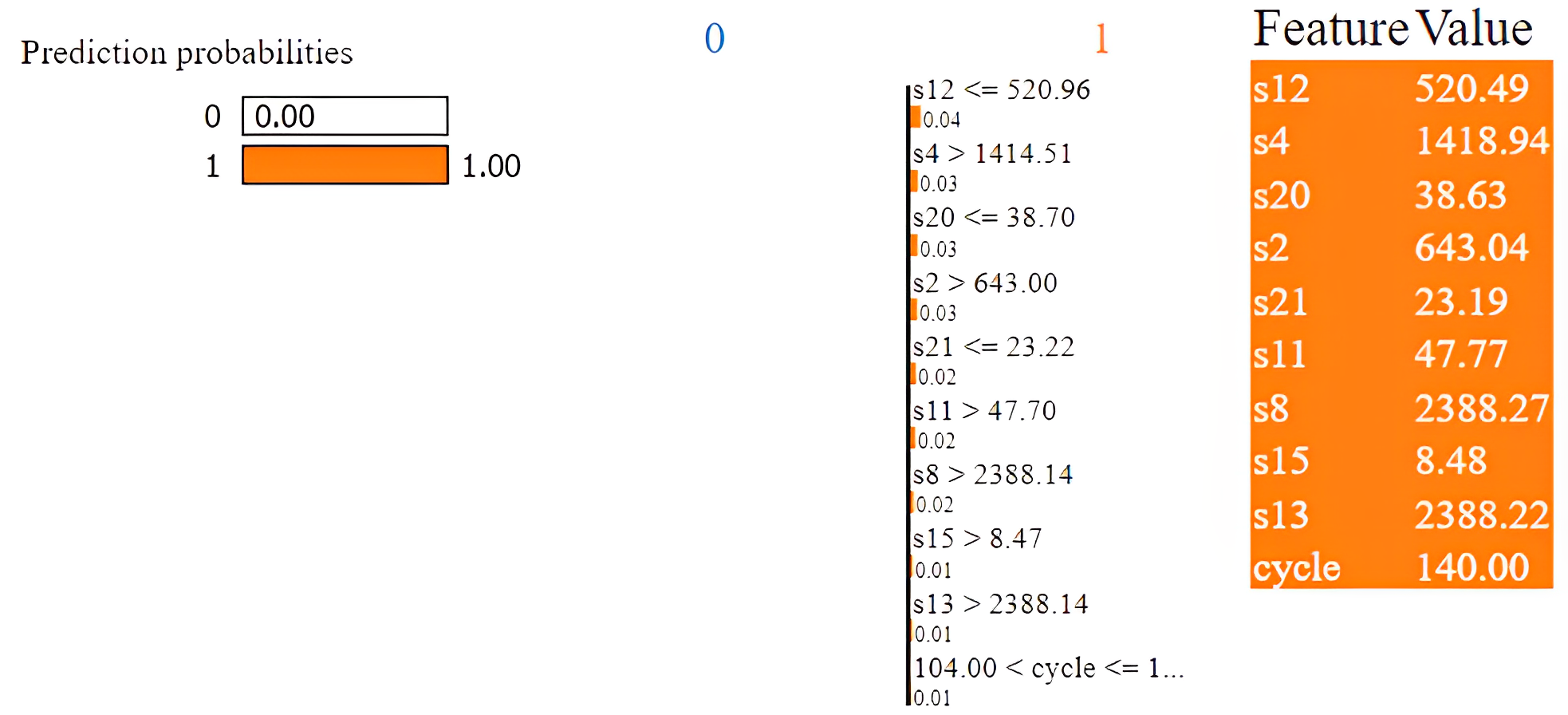}
    \caption{True Negative Case of Naive Bayes Classifier}
    \label{fig:samp4_nb_final}
\end{figure}

\begin{figure}[h]
    \centering
    \includegraphics[width=0.65\textwidth]{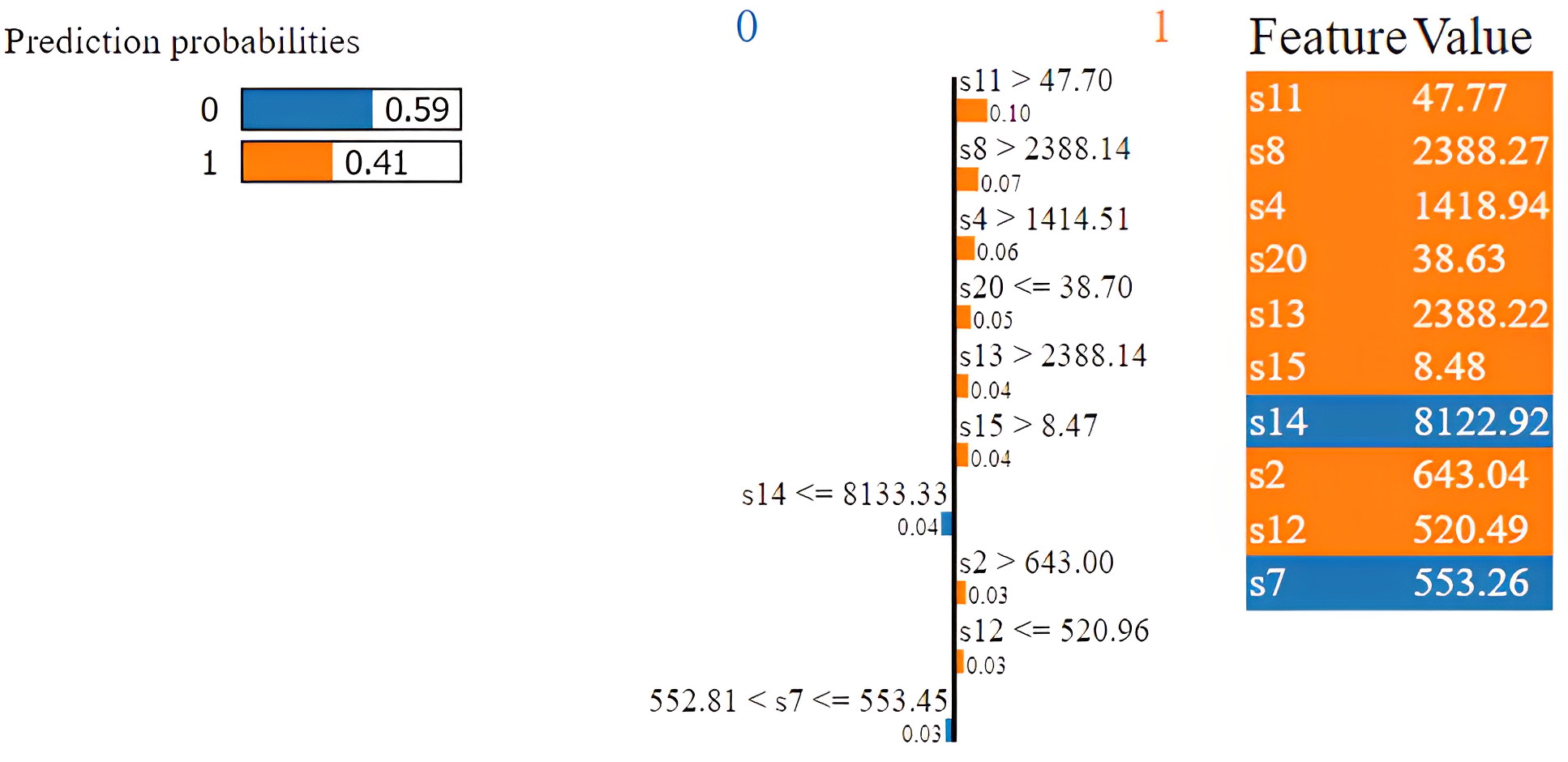}
    \caption{False Positive Case of Gradient Boosting Classifier}
    \label{fig:samp4_gradboost_final}
\end{figure}

On the other hand, in Fig. \ref{fig:samp4_nb_final} the Naive Bayes model properly identified the sample as 1 in this situation. The sensors' s12, s4, s20, s2, s21, s11, s8, s15, and s13 results are what allow it to be accurately classified. After applying Lime to the Gradient Boosting model for Sample 4 in Fig. \ref{fig:samp4_gradboost_final}, the model misclassified the data. Due to the higher values of s14 and s7 in class 0, the confidence level is only 41\%.

It is clear from the lime implementation for sample 6 that the value of s4 is what caused the sample's misclassification. For sample 4, the value of s7 increases the probability of its misclassification.
\\
Due to the fact that the final processed data for this time series dataset are merged, we were unable to use lime on LSTM, GRU, or RNN. In order to understand why machine learning models didn't perform well than the deep learning models, we applied lime to the models. Although it is known that Lime is only relevant for text and feature-based data, this is not how our final processed data are. An example of the outcome from the processed data is shown in Table \ref{tab:table5}.

\begin{table}[h!]
\label{tab:table5}
\caption{Final output of processed data}
\centering
\begin{tabular}{cccccc}
{[}{[}0.15512465 & 0.55172414 & 0.25       & ... 0. & 0.5503876  & 0.6974593{]}  \\
{[}0.15789473 & 0.39655173 & 0.5833333  & ... 0. & 0.37209302 & 0.59748685{]} \\
{[}0.16066483 & 0.5344828  & 0.16666667 & ... 0. & 0.51937985 & 0.602596{]}   \\
\multicolumn{6}{c}{... ... ...}                                         \\
{[}0.28531855 & 0.43103448 & 0.33333334 & ... 0. & 0.56589144 & 0.49461475{]} \\
{[}0.28808865 & 0.33908045 & 0.25       & ... 0. & 0.4108527  & 0.5223695{]}  \\
{[}0.29085872 & 0.5689655  & 0.8333333  & ... 0. & 0.34108528 & 0.5024855{]}{]}
\end{tabular}
\end{table}



\subsection{Performance Comparison}

\begin{table}[h!]
\caption{Comparison of our proposed models with some previous work}
\begin{tabular}{ c|c|c } \hline
\textbf{Reference Paper} &  \textbf{Performance} 
                    & \textbf{Our Proposed Model}\\  \hline
                    Adryan Fitra Azyus et al.\cite{Ref3} & Random Forest (90.3\%) &  Random Forest (96.1\%) \\ \hline
                    Srikanth Namudur et al.\cite{Ref5} & LSTM (87.3\%) & LSTM (96.42\%) \\ \hline
                    Abdeltif Boujamza et al.\cite{Ref6} & LSTM (96\%) & LSTM (96.42\%) \\ \hline
                    \end{tabular}
                    \label{tab:comp}
\end{table}

A comparison of our proposed models with some Table \ref{tab:comp} shows a comparison of our suggested approaches with some previously done work. The table clearly shows that the models we've provided are more accurate than those in past studies, and also further illustrates how the use of lime distinguishes our work from other studies in the earlier part.

\section{Conclusion and Future Work}In this paper, LSTM, Bi-LSTM, RNN, Bi-RNN, and GRU with some machine learning models such as Random Forest, KNN, Naive Bayes and Gradient boosting are used to apply the sensor's information to predict aircraft engine failure in a specified number of cycles. Using a simple scenario and only one data source (sensor readings), we have shown how deep learning and machine learning can be used to make predictions in predictive maintenance and discovered that deep learning models beat machine learning models, which has been proven by using the implementation of lime. When it comes to sequential time series forecasting, LSTM has an advantage over competing models in that it can take real-time data and apply it. Furthermore, LSTM has the ability to avoid issues with long-term short-term reliance. GRU, on the other hand, outperforms competitors because it makes recurrent neural networks more memory-efficient and makes model training simpler.
In the future, the Internet of Things (IoT) can be used to collect real-time data from aircraft engines and transmit it to maintenance personnel for analysis. This research, in our opinion, will be useful in the aviation industry.

%
%

\end{document}